\documentclass{article}
\usepackage{arxiv}

\usepackage[utf8]{inputenc} 
\usepackage[T1]{fontenc}    
\usepackage{hyperref}       
\usepackage{url}            
\usepackage{booktabs}       
\usepackage{amsfonts}       
\usepackage{nicefrac}       
\usepackage{microtype}      
\usepackage{lipsum}
\usepackage{float}
\usepackage{graphicx}
\usepackage{multirow}
\usepackage{array}
\usepackage{comment}
\usepackage{longtable}
\usepackage{ragged2e}
\usepackage{subfig}

\title{Machine Learning-Based Heart Disease Diagnosis: A Systematic Literature Review}

\author{
  Md Manjurul Ahsan \\
  Industrial and Systems Engineering\\
  University of Oklahoma\\
  Norman, Oklahoma-73071 \\
  \texttt{ahsan@ou.edu} \\
   \And
 Zahed Siddique \\
  Department of Aerospace and Mechanical Engineering\\
  University of Oklahoma\\
  Norman, Oklahoma-73071\\
  \texttt{zsiddique@ou.edu}} 

\begin{document}
\maketitle

\begin{abstract}
Heart disease is one of the significant challenges in today’s world and one of the leading causes of many deaths worldwide. Recent advancement of machine learning (ML) application demonstrates that using electrocardiogram (ECG) and patients’ data, detecting heart disease during the early stage is feasible. However, both ECG and patients’ data are often imbalanced, which ultimately raises a challenge for the traditional ML to perform unbiasedly. Over the years, several data level and algorithm level solutions have been exposed by many researchers and practitioners. To provide a broader view of the existing literature, this study takes a systematic literature review (SLR) approach to uncover the challenges associated with imbalanced data in heart diseases predictions. Before that, we conducted a meta-analysis using 451 reference literature acquired from the reputed journals between 2012 and November 15, 2021. For in-depth analysis, 49 referenced literature has been considered and studied, taking into account the following factors: heart disease type, algorithms, applications, and solutions. Our SLR study revealed that the current approaches encounter various open problems/issues when dealing with imbalanced data, eventually hindering their practical applicability and functionality.
\end{abstract}


\section{Introduction}
Cardiovascular diseases (CVD) are one of the leading causes of death globally~\cite{ahsan2021effect}. Some of the most known CVD includes coronary heart disease, cerebrovascular disease, peripheral arterial disease, rheumatic heart disease, and congenital heart disease. According to the world health organization (WHO), around 17.9 million people die every year due to heart-related complications and heart disease. More than four out of five CVD deaths are due to heart attack and strokes. Unhealthy diets, low physical activity, alcohol abuse, and tobacco use are some of the potential risk factors that accelerate heart-related complications. As a result, intermediate-risk factors such as blood pressures, high blood glucose levels, extreme blood lipids, overweight, and obesity are observed among the individuals~\cite{WHO}. However, identifying those at high risk of CVD at the early stages and providing appropriate treatments can prevent unexpected and premature deaths.\\
More than three-quarters of all deaths from cardiovascular disease (CVD) are in low- and middle-income nations~\cite{WHO}. With the growing population and number of heart disease patients, it is becoming more challenging to provide individuals affordable diagnoses in a less developed country like Bangladesh, India, and some African nations, where proper screening procedures for patients with heart disease symptoms are still questionable due to financial crisis, less access to adequate and equitable health care equipment and facilities~\cite{lip2017antithrombotic,islam2013coronary}. Additionally, with existing facilities, it is not affordable for the general people to take the opportunity to diagnosis heart disease.\\
 Electrocardiogram (ECG) is employed to diagnose CVD. However, visually identifying long-term ECG abnormalities takes time and effort. With the advent of machine learning (ML) applications in the medical domain, many researchers and practitioners found machine learning-based heart disease diagnosis (MLBHDD) systems as cheap and flexible approaches~\cite{nahar2013computational,sree2012cardiac,liu2012intelligent}. As a consequence, several studies proposed MLBHDD using different heart disease datasets~\cite{nahar2013computational,exarchos2014multiscale,wiharto2016intelligence}. For instance, Bashir et al. (2016) used various machine learning approaches such as Naive Bayes (NB), Decision Tree (DT) based on Gini index, DT based information gain, instance-based learner, and Support Vector Machines (SVM) to develop an ensemble-based model to focus on prediction and analysis of heart patients and achieved an accuracy of 87.37\%~\cite{bashir2016multicriteria}; Daraei and Hamidi (2017) presented an ML-based Myocardial infarction (MI) prediction model using J48 algorithms and reported 82.57\% accuracy~\cite{daraei2017efficient}.
Recently, Deep Learning (DL) added an additional layer and demonstrated the benefit of developing data-driven heart disease diagnosis approaches with an accuracy close to 100\%. A Convolutional Neural Network (CNN) based coronary heart disease diagnosis model by Dutta (2020)~\cite{dutta2020efficient}, Deep Neural Network (DNN) based model, named CraftNet by Li et al. (2020)~\cite{li2020craftnet} are some of the examples out of many proposed DL based heart disease diagnosis model.\\
One of the potential drawbacks of machine learning (ML) and the DL-based solution is that in many cases, there is no proper explanation of the models' behaviors during the final predictions~\cite{ahsan2021detection}. For instance, a Deep Learning-based model contains several hidden layers, but it is difficult to interpret how each layer contributes during the final prediction, in most cases~\cite{ahsan2021detecting}. Apart from this, the biased performance of ML algorithms towards the majority class is another potential challenge. A majority class occurs when a data set contains maximum value in one class compared to other classes, also known as the imbalanced dataset~\cite{fernandez2011addressing}. Therefore, questions remain regarding the ML-based models' performance in terms of biasness, fairness, and interpretability~\cite{ahsan2021detection}. Thus, there is a need to identify the recent trend, techniques, gaps, and future opportunities related to ML-based heart disease diagnosis.\\
Table~\ref{tab1} presents an overview of some of the previously published literature reviews on heart disease diagnosis using ML approaches. From Table~\ref{tab1}, it can be observed that most of the referenced literature emphasizes machine learning approaches while the systematic literature review (SLR) is mostly ignored.  For instance, Benhar et al. (2020) published an SLR whose primary concern was different ML algorithms and data preprocessing techniques used in heart disease diagnosis~\cite{benhar2020data}. However, most of the existing heart disease dataset contains imbalanced data; therefore, the performance of ML on imbalanced heart disease data is also required to be analyzed; some of the studies conducted the traditional literature review that did not follow the proper SLR process~\cite{rath2021exhaustive,hoodbhoy2021diagnostic}. Additionally, many studies did not provide the time periods of the considered literature~\cite{verma2021effective,kumar2021machine}. Therefore, with the rising of ML-based diagnosis, it is necessary to conduct an SLR that may bridge the gap between existing surveys by providing SLR with meta-analysis.
\begin{table*}[h!]
  \caption{Related research}
    \centering\resizebox{\textwidth}{!}{
    \begin{tabular}{@{}p{.3\linewidth}p{.08\linewidth}p{.25\linewidth}p{.1\linewidth}p{.08\linewidth}p{.08\linewidth}p{.05\linewidth}p{.05\linewidth}p{.05\linewidth}@{}}\toprule
         Paper title&	Time period&	Study focus&	Algorithms&	Imbalance challenges&	Evaluation metrics&	Meta analysis&	Content analysis&	SLR\\\midrule
       Data preprocessing for heart disease classification: A systematic literature review~\cite{benhar2020data} & 2000-2019& Data preprocessing& \checkmark& & \checkmark& \checkmark&\checkmark&\checkmark\\ 
         Effective prediction of heart disease
         using data mining and machine learning: a review~\cite{verma2021effective}& Not specified & Data mining and Algorithm techniques& \checkmark&&\checkmark\\
         Machine learning based heart disease diagnosis using non-invasive methods: a review~\cite{kumar2021machine}& Not specified&	Features, samples, algorithms&\checkmark&	&\checkmark\\
         Diagnostic accuracy of machine learning models to identify congenital heart disease: a meta-analysis~\cite{hoodbhoy2021diagnostic}&Until march 31, 2020&Estimating diagnostic accuracy of machine learning models&\checkmark&&\checkmark&\checkmark&&\checkmark\\
         An exhaustive review of machine and deep learning-based diagnosis of heart disease~\cite{rath2021exhaustive}&Not specified& kinds of test data, characteristics, retrieved signals from patients, sources of standardized data sets, feature extraction and feature reduction techniques &\checkmark&&\checkmark&\checkmark \\
         Our study& 2012-2021& Machine learning application in heart disease diagnosis& \checkmark&\checkmark&\checkmark&\checkmark&\checkmark&\checkmark\\\bottomrule
    \end{tabular}}
  
    \label{tab1}
\end{table*}

In light of the rising number of articles published in MLBHDD, it is important to devise new insights and research directions based on the current body of knowledge. As a result, a systematic literature review (SLR) is carried out using 451 research papers that have been pooled from the Scopus database. Initially, collected 451 papers were used to conduct metadata analysis, and among them, 49 papers are being used for in-depth analysis. The fundamental goal of the metadata analysis is to address the following research questions: what are the top contributing countries, institutions, and subject areas? Who are the potential authors? How much research is incorporated with fundings? The in-depth analysis of 49 papers addressed the following research questions: What are the current machine learning and deep learning-based approach used in heart disease diagnosis? What are the current strategies to handle datasets containing an imbalance class ratio?\\
The end goal of the SLR is to serve as a reference point for both theorists and practitioners, not only providing an overview of recent trends and techniques but also by identifying the research gap that might help in developing an advanced MLBHDD model. The following section rest of the paper is organized as follows: in Section~\ref{methods}  the SLR methodology is briefly explained, followed by the observations and findings in Section~\ref{observation}. Finally, Section~\ref{discussion} summarize our findings, followed by the identification of opportunities for future work in Section~\ref{conclusions}.

\section{Methods}\label{methods}
A systematic literature review (SLR) is a review in which questions are formulated and systematic and explicit procedures are used to find, select, and critically appraise relevant research in order to gather and evaluate data from the studies included in the review~\cite{okoli2010guide}. This method is chosen because it provides an accurate and reliable manner to synthesize academic literature and is widely accepted in many research domains. The SLR is reported in accordance with the Preferred Reporting Information for Systematic Reviews and Meta-Analysis (PRISMA) recommendations. This study does not need ethical approval. Even though PRISMA is not a quality assessment approach, it is acceptable in the research domain due to its 27-evidence based checklist and four-phase analysis, ultimately providing the opportunity of clarity and transparency of any systematic literature reviews (SLR)~\cite{tricco2018prisma}. 
\subsection{Identification of the data}
\begin{figure}
     \centering
     \includegraphics[width=\textwidth]{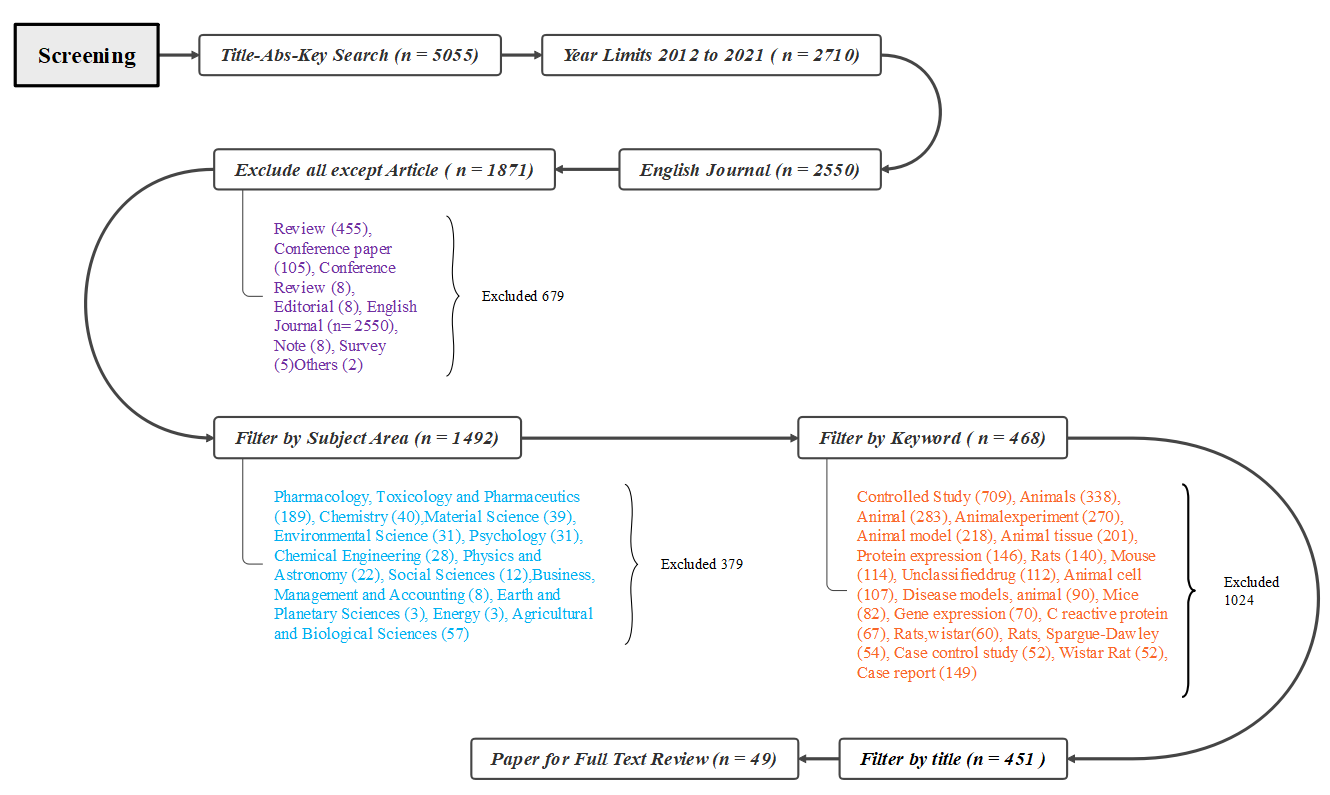}
     \caption{Flow diagram of the PRISMA approach utilized in this research}
     \label{fig:fig1}
 \end{figure}

A comprehensive search study was executed using Scopus integrated database, including all major publishers such as Emerald, Taylor and Francis, Springer, IEEE, and Willey. Scopus database has been considered a reliable database by many researchers to conduct SLR due to high-quality indexing contents~\cite{fahimnia2015green,malviya2015green}. The search spans from 2012 through November 15, 2021, and includes all relevant publications published within this period. When searching for relevant publications, we have used keywords like "heart," "machine learning," "imbalance," "diagnostic," and "deep learning. To broaden the search area, the Boolean operators are used along with different keywords.\\
The search method was created and implemented by one investigator (M.A.) with the help of another investigator (Z.S.). Controlled vocabulary and related keywords are strictly followed to narrow down the search radius appropriately. The overall search strategy is demonstrated in Fig.~\ref{fig:fig1}.
\subsection{Screening initial data and determining eligibility}
Using selected keywords, the initial search in the Scopus database displayed 5055 articles. Once the year limit was applied from 2012 to November 15, 2021, the number of articles was reduced to 2710. Given the restriction of document type, language, filter by subject area, filter by keywords limits the total number of articles to 468. After using keywords, search procedure, 468 articles were identified for the title and abstract screening. Resulted of 468 article information was imported by one investigator (Z.S.) as excel CSV data for future investigation. Duplicates were identified and removed using excel duplicate functions. Then the remaining unique 451 articles titles and abstracts were screened for further inclusion. Two reviewers (M.A. and Z.S.) independently screened 451 articles titles and abstracts and used a standard extraction form. Conflicts were resolved by discussion. The study that is not relevant to machine learning but related to heart disease or vice-versa was excluded.\\
Additionally, reviews, non-human studies, and book chapters were also excluded from this study. After title and abstract screening total of 49 articles were read in full-text, and all 49 articles met all inclusion criteria. Fig.~\ref{fig:fig1} presents the exclusion and inclusion procedures used during this study. There were several reasons for excluding the article for the full-text screening:
\begin{enumerate}
    \item Not containing imbalance data analysis methods
    \item Only focused on models’ performance instead of the possible limitations
    \item Not a peer review journal article
    \item Full-text inaccessibility
\end{enumerate}
\section{Observations and findings}\label{observation}
The results of the metadata analysis and insights are presented in the following section. The findings were based on a content analysis of 49 publications and a metadata study of 451 selected papers.
\subsection{Metadata analysis}
The metadata study included 451 papers organized by year, journals, authors, countries, subject areas, funding, and institution.
\subsubsection{Publication by year}
\begin{figure*}
\
    \includegraphics{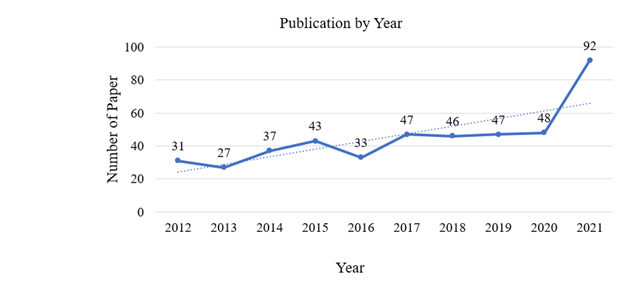}
    \caption{Publications of ML-based heart disease diagnosis (MLBHDD) by year}
    \label{fig:fig2}
\end{figure*}

Fig.~\ref{fig:fig2} shows the number of papers published on heart disease using ML approaches over the past ten years based on our selected 451 papers. From 2012 to 2020, the steady growth of the publication is observed, while there is an exponential growth in 2021. For instance, the number of papers published in 2020 was around 48, while in 2021, it is 92. Moreover, over time it can be seen that the importance of imbalance classification problem in heart disease diagnosis gets much attention. The number of papers published in 2021 is significantly higher than in any previous year. As a result, increased attention and concern are being dedicated to heart disease diagnosis with imbalance classification problems, parallel with other data-driven challenges.
\subsubsection{Scholarly journal articles published between 2012 and 2021}
\begin{figure*}
    \centering
    \includegraphics{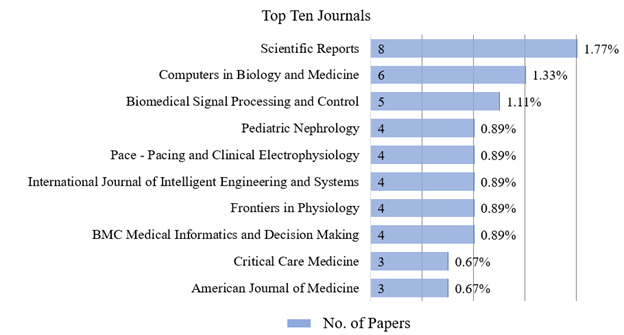}
    \caption{Publications by Journals}
    \label{fig:fig3}
\end{figure*}

Fig.~\ref{fig:fig3} indicates that of the 451 publications, the most were published in Scientific Reports, with eight. This equates to 1.77 percent of the total. However, considering ten years of publications, MLBHDD papers in scientific Report is relatively low. For instance, among 451 papers, the number of published articles for Journal of Critical Care Medicine and International Journal of Intelligent Engineering and Systems are just 3, even though they are the top ten journals in terms of a maximum number of papers published. 
\\Thus, it can be assumed that publications by journals are very diverse, and no single Journal dominates this field in terms of published articles in heart disease diagnosis. Among all the Journals, the Critical Care Medicine Journal has the highest impact factors (7.598), and Scientific Report is the second-best Journal with impact factors of 5.133, as shown in Table~\ref{tab:tab2}.
\begin{table}
   \caption{Top ten Journals with impact factor}
    \centering
    \begin{tabular}{@{}ll@{}}\toprule
       Journals&	Impact \\&factor (2021)\\\midrule
Scientific Reports&	5.133\\
Computers in Biology and Medicine&	4.589\\
Biomedical Signal Processing and Control&	3.137\\
Pediatric Nephrology&	3.174\\
Pace - Pacing and Clinical Electrophysiology&	1.156\\
International Journal of Intelligent\\ Engineering and Systems&	1.17\\
Frontiers in Physiology&	4.134\\
BMC Medical Informatics\\
and Decision Making&	2.796\\
Critical Care Medicine&	7.598\\
American Journal of Medicine&	4.76\\\bottomrule

    \end{tabular}
 
    \label{tab:tab2}
\end{table}
\subsubsection{Publication by authors}
\begin{figure*}
    \centering
    \includegraphics{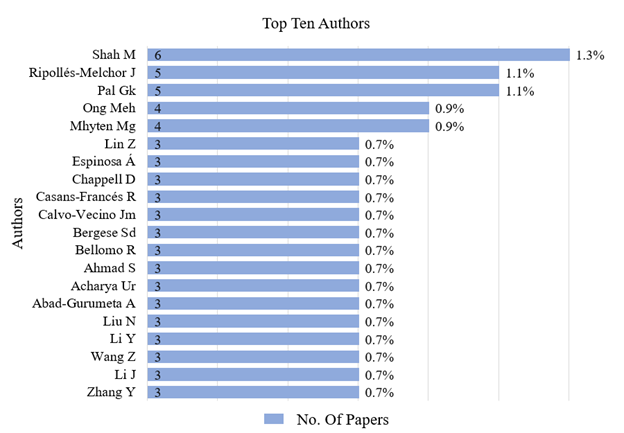}
    \caption{Top 10 authors in heart disease analysis}
    \label{fig:fig4}
\end{figure*}

According to Fig.~\ref{fig:fig4}, Mahek Shah has published the most publications on MLBHDD (6 out of 451). Simultaneously, Javier Ripollés-Melchor and Gopal Krushna Pal ranked second by publishing five articles. 14 authors, on the other hand, published three papers over the years and were all listed among the top ten authors. Despite this, the number of individual published articles is relatively low when compared to the ten years of analysis.
\subsubsection{Publication by citations}
More often, citation of the papers also provides a schematic view of the influential authors in the relevant domains. Table~\ref{tab:tab3} presents the ten most cited articles in the Scopus database until November 2021. However, the number of citations may differ slightly with google scholar due to their different indexing procedure and time. From Table~\ref{tab:tab3}, it can be observed that the paper published by Acharya et al. in 2017 received the maximum number of citations (448) and 89.6 citations per year~\cite{acharya2017deep}. Table~\ref{tab:tab3} shows that all of the authors referred between 28 and 448 times that considered MLBHDD between 2012 to 2021.
\begin{table*}
 \caption{Top 10 cited papers published in MLBHD between 2012-2021}
    \centering\resizebox{\textwidth}{!}{
    \begin{tabular}{llll}\toprule
         Author&	Paper title&	Total citations&	Total citation/year\\\midrule
         Acharya et al. (2017)~\cite{acharya2017deep}&	A deep convolutional neural network model to classify heartbeats&
	448&
	89.6\\
	Nahar et al. (2013)~\cite{nahar2013computational}&	Computational intelligence for heart disease diagnosis:\\& a medical knowledge driven approach&
	155&	17.222\\
	Plawiak et al. (2020)~\cite{plawiak2020novel}&	Novel deep genetic ensemble of classifiers for\\& arrhythmia detection using ECG signals
	&103&	51.5\\
	Novikov et al. (2018)~\cite{novikov2018fully}&	Fully Convolutional architectures for\\& multiclass segmentation in chest radiographs&
	92&	23\\
	Rajesh et al. (2018)~\cite{rajesh2018classification}&	Classification of imbalanced ECG beats using\\& re-sampling techniques and AdaBoost ensemble classifier&
	63&	7\\
Yang et al. (2018)~\cite{yang2018automatic}&	Automatic recognition of arrhythmia based on\\& principal component analysis network\\& and linear support vector machine&
	61&	15.25\\
Sellami et al. ( 2019)~\cite{sellami2019robust}&	A robust deep convolutional neural network\\& with batch-weighted loss for\\& heartbeat classification
	&57&	19\\
Awad et al. (2019)~\cite{awad2017early}&	Early hospital mortality prediction of\\& intensive care unit patients using an \\&ensemble learning approach
	&51&	10.2\\
Sakr et al. (2017)~\cite{sakr2017comparison}&	Comparison of machine learning techniques\\& to predict all-cause mortality using fitness\\& data: the Henry ford Exercise testing (FIT) project&
	32&	6.40\\
Awan et al. (2012)~\cite{awan2019machine}&	Machine learning-based prediction\\& of heart failure readmission or death:\\& implications of choosing the right model\\& and the right metrics&
	28	&9.33\\\bottomrule
    \end{tabular}}
   
    \label{tab:tab3}
\end{table*}
\subsubsection{Publication by countries}
\begin{figure*}
    \centering
    \includegraphics{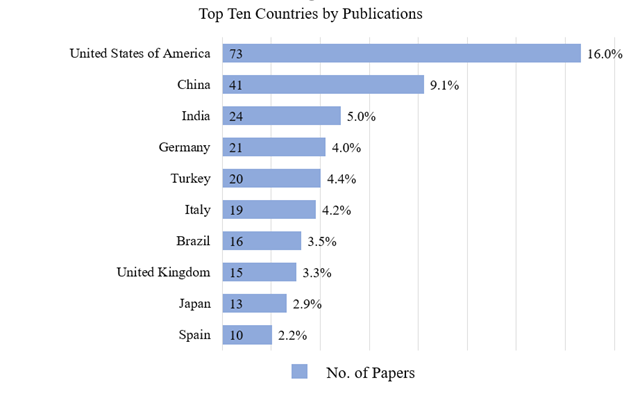}
    \caption{Top ten countries that contributed to MLBHDD literature}
    \label{fig:fig5}
\end{figure*}

Fig.~\ref{fig:fig5} showed that the United States published the most publications in MLBHDD, up to 16\% of the total 451 referenced literature. Respectively, China and India are placed 2nd and 3rd by publishing 41 (9.1\%) and 24 (5.0\%) papers. It has become clear that the USA dominates in this research field by publishing 16\% of the papers while Asian countries like China, India, and Japan’s combined publication is around 17\%. Thus, it could be concluded that researchers and practitioners in the USA emphasized more on MLBHDD related papers than any other nation.
\subsubsection{Most frequently words used in the titles and keywords}
Table~\ref{tab:tab4} presents the most frequent single, double and triple keywords that appeared in the titles of the papers. The most frequently used keywords are determined using the R-software tool. Even though our primary objective was to identify and analyze the article that solely focused on the terms like machine learning, deep learning, imbalance class, heart disease, surprisingly, we found that for single keywords, only "heart" and "disease" placed among the top most used words by the authors in the titles. For instance, "heart" is the most used single keyword (115), "heart failure" (38) as double keywords, and "heart rate variability" (17) as triple keywords have appeared as the most frequently used keywords in the titles.
\begin{table*}
\caption{Most frequent single, double and triple keywords from the titles}
    \centering
    \begin{tabular}{llllll}\toprule
         Unigrams	&Frequency&	Bigrams&	Frequency&	Trigrams&	Frequency\\\midrule
         
         Heart&	115&	Heart failure&	38&	Heart rate variability&	17\\
         Patients&	98&	Heart rate&	25&	Coronary artery disease&	7\\
Cardiac&	87&	Heart disease&	20&	Acute kidney injury&	6\\
Disease&	40&	Myocardial infarction&	17&	Chronic heart failure&	6\\
Failure&	40&	Rate variability&	17&	Coronary heart disease&	6\\
Study&	38&	Cardiac Surgery&	15&	Heart failure patients&	6\\
Autonomic&	32&	Fluid overload&	12&	Intensive care unit&	6\\
Surgery&	32&	Left ventricular&	12&	Convolutional neural network&	5\\
Learning&	28&	Machine learning&	12&	Heart disease prediction&	5\\
Cardiovascular&	27&	Coronary artery&	11&	Acute myocardial infarction&	4\\\bottomrule

    \end{tabular}
    
    \label{tab:tab4}
\end{table*}
\\
However, we found interesting results when we focused on authors' keywords in keyword sections of the papers. Table~\ref{tab:tab5} shows the most frequently used words in keyword sections of the papers. "heart failure" and "heart rate variability" is the most common word used in keywords (35 times), followed by "machine learning" (27) and "deep learning" (17), which are 2nd and 3rd most common words chosen by the authors in the keyword sections.
\begin{table}
 \caption{Most frequently utilized words in keyword sections}
    \centering
    \begin{tabular}{ll}\toprule
       Terms&	Frequency\\\midrule
Heart failure&	35\\
Heart rate variability&	35\\
Machine learning&	27\\
Deep learning&	17\\
Myocardial infarction&	13\\
Classification&	12\\
Cardiopulmonary bypass&	10\\
Mortality&	10\\
Stroke&	10\\
Autonomic nervous system&	9\\\bottomrule

    \end{tabular}
   
    \label{tab:tab5}
\end{table}

\begin{figure*}
    \centering
    \includegraphics{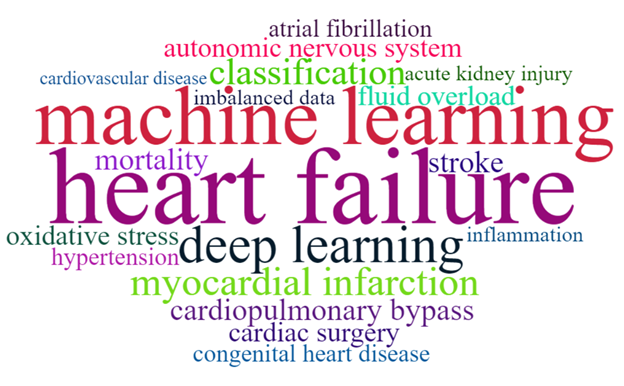}
    \caption{Word cloud for most frequently used keywords in MLBHDD publications}
    \label{fig:fig6}
\end{figure*}

 To find the most prevalent terms in a complicated setting, a word cloud is a simple way to detect the common themes and keywords that are utilized in the referenced articles. Software generated word clouds are shown in Fig.~\ref{fig:fig6}, with larger and bolder fonts showing the most often used words and smaller and more common fonts highlighting the less frequently used phrases.
\subsubsection{Publication by institutions}
\begin{figure*}
    \centering
    \includegraphics{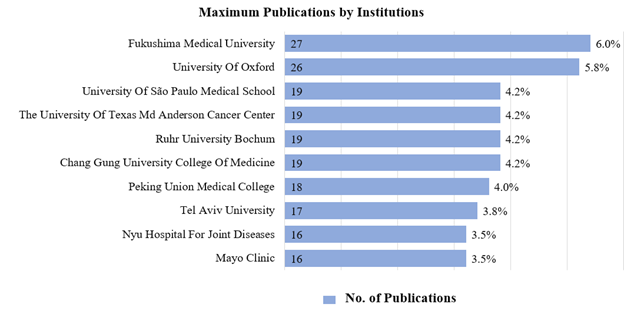}
    \caption{Top ten institutions based on number of publications}
    \label{fig:fig7}
\end{figure*}

Fig.~\ref{fig:fig7} illustrates leading publications by author’s affiliation. The Figure shows that Fukushima Medical University, Japan published the highest number of papers in MLBHDD literature. The institutes published 27 articles, approximately 6\% of the total 451 published papers. University of Oxford, England, is the second institution with the highest number of papers published, followed by the University of São Paulo Medical School, Brazil. It is interesting to observe that, even though the USA is the top country considering the publications, USA-based University, The University of Texas Md Anderson cancer center institute placed 4th. On the other hand, while Japan placed 9th out of the top ten countries in terms of institutions, Japan-based University, Fukushima Medical University, positioned 1st. Thus, it could be assumed that the authors from Fukushima Medical University are more active than any other institution in terms of collaborative works.
\subsubsection{Contribution by subject area}
\begin{figure*}
    \centering
    \includegraphics{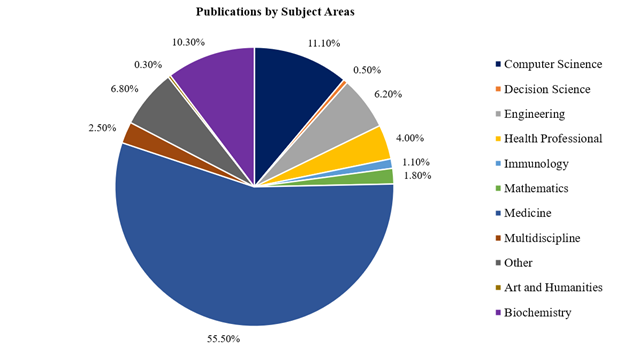}
    \caption{Maximum number of publications based on subject areas}
    \label{fig:fig8}
\end{figure*}

Fig.~\ref{fig:fig8} demonstrates that multiple domains published heart disease diagnosis-related literature. However, Medicine is the most dominated subject area where pooled literature is around 55\%. The second most researched subject is computer science (11.10\%), followed by biochemistry (10.30\%). Apart from this, other disciplines such as engineering, health professionals, and decision sciences also published several research papers, indicating the growing interest in heart disease-related research in multi-discipline.
\subsubsection{Publication by funding}
\begin{figure*}
    \centering
    \includegraphics{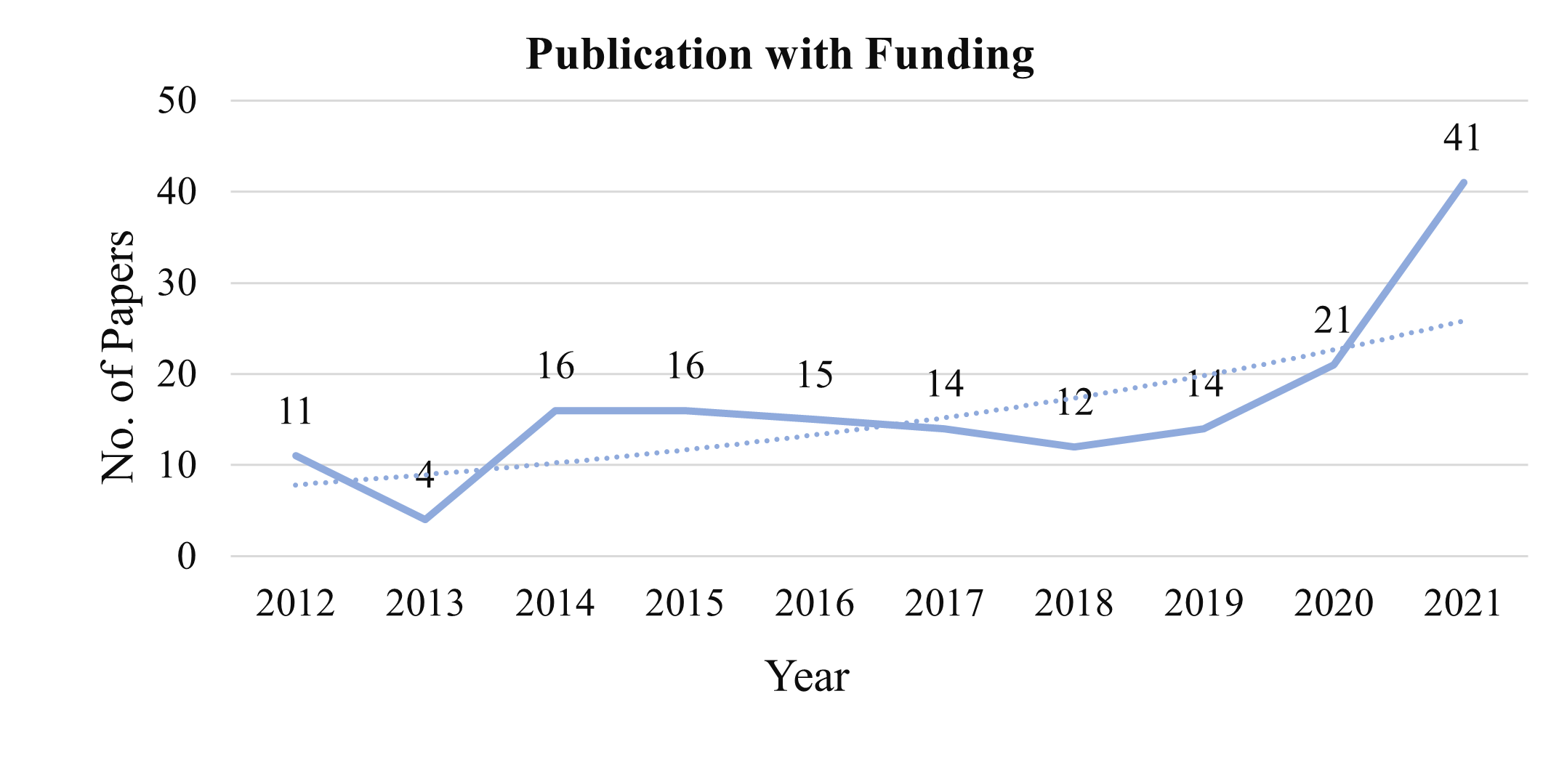}
    \caption{Number of papers incorporated with funding sources over the times}
    \label{fig:fig9}
\end{figure*}

Fig.~\ref{fig:fig9} demonstrates the number of research papers related to funded projects. The number of articles published between 2012 to 2018 fluctuated in terms of funding sources. However, from 2019 to 2021, exponential growth is observed, indicating that the MLBHDD draws attention to researchers, practitioners, and the funder, ultimately reflected by the increase of funded and published research.
\subsection{Insights of MLBHDD}
This section examines 49 studies that focus on imbalanced data, heart disease, and machine learning. The objective is to conduct an in-depth analysis of those 49 publications in terms of concept, technique, trends, and future scopes that may provide insights to theorists and practitioners.
\subsubsection{Disease types}
\begin{figure*}
    \centering
    \includegraphics{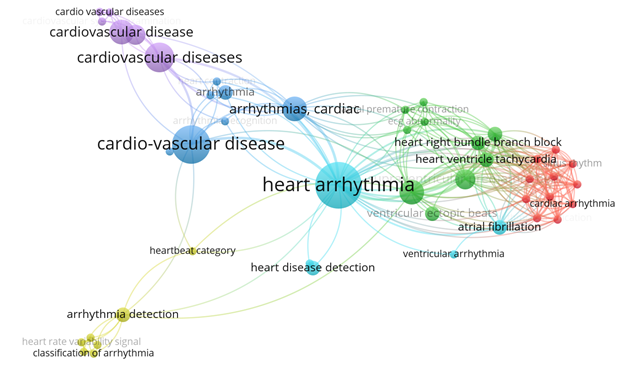}
    \caption{Illustration of most frequently heart-related disease reported by the reference literature (developed by VOSviewer software)}
    \label{fig:fig10}
\end{figure*}

With the advancement of ML-based approaches, researchers and practitioners adopted data-driven approaches to diagnose heart-related disease using electrocardiogram (ECG) signals. However, the patients face problems like delayed diagnosis due to the undergo of different routine tests and consult a doctor only after the symptoms become severe. In contrast, ML-based approaches allow early-stage diagnosis, which can be conducted by the subject himself on a routine basis utilizing low-cost and compact sensors~\cite{bashir2016multicriteria}.\\
Heart disease can be identified at an early stage by analyzing heartbeat rhythms. Heartbeat can be divided into five categories: non-ectopic, supraventricular ectopic, ventricular ectopic, fusion, and unknown beats. The occurrence of abnormal heartbeat is also known as arrhythmia. Among the selected 49 literature, at least 13 papers considered arrhythmia due to its fatal consequences. It is one of the primary causes of morbidity and mortality among cardiac patients~\cite{yang2018automatic}. Therefore, early diagnosis is essential to provide adequate treatment and medication for patients suffering from cardiac arrhythmia. For instance, Yang et al. (2018), developed a novel heart disease recognition method using a Linear Support Vector machine to detect arrhythmia and achieved around 97.77\% accuracy for the imbalance data and 97.08\% accuracy for the noise-free ECGs~\cite{yang2018automatic}. Romdhane et al. (2020) proposed a CNN-based heartbeat segmentation approach to identify arrhythmia and achieved an accuracy of 98.41\%~\cite{romdhane2020electrocardiogram}. Both studies were carried out using the MIT-BIH arrhythmia heart disease open repository dataset. On the other hand, Che et al. (2021) utilized a CNN-based approach to extract ECG signal temporal information using real-world data~\cite{che2021constrained}.\\
Most of the researcher uses the tabular dataset to detect different types of heart disease using machine learning approaches. Instead of specifying the disease, most authors used heart disease as a general term. For instance, Nahar et al. (2013)~\cite{nahar2013computational} and Gan et al. (2020)~\cite{gan2020integrating} both studies reported their result considering heart disease using the Cleveland dataset. The dataset contains 13 attributes to detect whether the patients have heart disease or not instead of any specific heart disease.\\
Fig.~\ref{fig:fig10} illustrates the most frequently heart-related disease reported by the referenced literature. From the Figure, it can be seen that “heart arrhythmia” and “cardiovascular disease” is the biggest cluster considering any other heart-related disease in terms of the number of papers. For instance, Minou et al. (2020) developed a classification technique using Random Forest (RF) and Decision Trees (DT) to detect cardiovascular disease~\cite{minou2020classification}. Kumar and Ramana (2021) proposed NN-based approaches to diagnosis the prognosis and severity of patients with cardiovascular disease~\cite{kumar2021cardiovascular}. Apart from this, some other studies reported cardiac arrest~\cite{baral2021novel,liu2012intelligent}, coronary heart disease~\cite{wiharto2016intelligence,dutta2020efficient}, and Myocardial infarction (MI)~\cite{wiharto2016intelligence,sharma2020myocardial} taking into account the MLBHDD approaches.
\subsubsection{Machine learning algorithms}
\begin{figure*}
    \centering
    \includegraphics{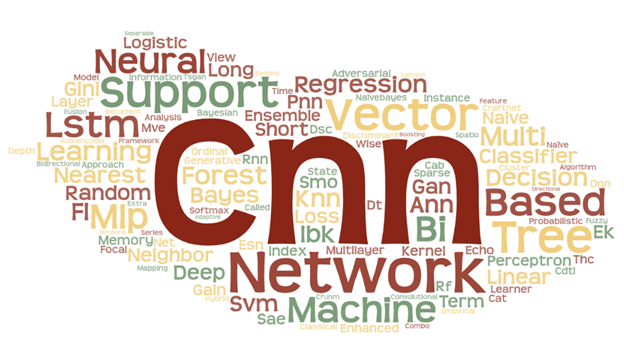}
    \caption{Most frequent Machine learning algorithms used in referenced literature}
    \label{fig:fig11}
\end{figure*}

 From Table~\ref{tab:tab6}, it can be easily understood that, over the years, researchers and practitioners displayed more interest on DL algorithms instead of traditional ML in developing MLBHDD models. Among 49 studies, at least 27 adopted a DL-based approach to developing a heart disease diagnosis model. For instance, Awan et al. (2019) used Multi-layer Perceptron (MLP) to predict heart failure patients' readmission within 30 days. Preliminary computational results show that, the proposed model can detect the heart failure patients with 48\% sensitivity and 70\% specificity~\cite{awan2019machine}. Li et al. (2020) proposed a Deep Neural Network (DNN)-based model named craftNet to accurately recognize the handcraft features to detect cardiovascular disease and achieved accuracy between 86.82\% to 89.25\%~\cite{li2020craftnet}. Dixit and Kala (2021) introduced 1D CNN model in order to detect heart disease patients in the early stage using cost-effective and compact ECG sensor. Their primary result shows that, with 300 actual patients' data, model is capable of detecting heart disease patients 93\% of the time accurately~\cite{dixit2021early}.\\
The second most adopted algorithm by the researcher is Support Vector Machines (SVM). For instance, Liu et al. (2012) used SVM to develop intelligent scoring systems for predicting cardiac arrest within 72 hours~\cite{liu2012intelligent}. Shah et al. (2020) tested SVM along with Random Forest (RF), Ordinal Regression, Logistic Regression (LR), and NB on the Cleveland dataset to detect heart disease patients. Among all algorithms, SVM demonstrated the best performance by achieving an accuracy of 95\%~\cite{shah2020heart}. Apart from SVM and CNN based approaches, other algorithms such as ensemble learning~\cite{wang2020left}, k-Nearest Neighbors (kNN)~\cite{sharma2020myocardial}, DT~\cite{minou2020classification}, linear Discriminant Analysis (LDA)~\cite{polat2018similarity}, Bayesian Networks (BN)~\cite{exarchos2014multiscale} are quite a few among many ML-based algorithms that are adopted by the researchers in developing MLBHDD model.\\
However, recently, many researchers show that, for balance and imbalance datasets, in both cases, Generative Adversarial Network (GAN) performed better than any existing methods. In consequence, some of the published literature in 2021 proposed GAN based model~\cite{wang2021cab,rath2021exhaustive,puspitasari2021generative}. For instance, Wang et al. (2021) proposed a novel GAN-based approach called CAB that can handle imbalance-related problems, and classification accuracy is around 99.71\% for arrhythmia patients~\cite{wang2021cab}. Rath et al. (2021) proposed a model combination of Long Short Term Memory (LSTM) and GAN that can accurately detect heart disease patients from the MIT-BIH dataset up to 99.4\%~\cite{rath2021exhaustive}. Fig.~\ref{fig:fig11} displayed the most used algorithms based on selected 49 referenced literature. The Figure shows that CNN is the most widely adopted algorithm by researchers and practitioners, followed by SVM.
\subsubsection{Imbalance challenges}
\begin{figure*}
    \centering
    \includegraphics{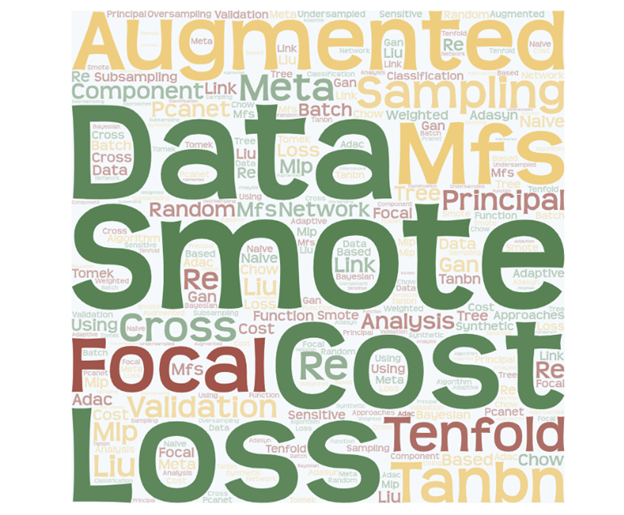}
    \caption{Most frequent approaches to solving imbalanced class problems, adopted by the referenced literature}
    \label{fig:fig12}
\end{figure*}

At the beginning of the study, one of the main goals was to identify the heart disease-related literature that considered imbalance data. However, after screening through most of the articles, it was observed that, the maximum study either uses actual data or adopted data from different open sources and in both cases, datasets were imbalanced. Therefore, during the quality assessment, it was identified that all 49 articles contained experimental results that are derived from the imbalanced dataset. However, not all of the studies deemed imbalanced data as a potential challenge. Most of the earlier studies were unaware of the impact of imbalanced data on model performances, which was addressed by many recent referenced literature. In that sense, different authors use different approaches to handle the imbalance issues. For instance, while Exarchos et al. (2015)~\cite{exarchos2014multiscale}, Bashir et al. (2016)~\cite{bashir2016multicriteria}, and Acharya et al. (2017)~\cite{acharya2017deep} rely on data level solutions, studies conducted by Yang et al. (2018)~\cite{yang2018automatic}, Sellami and Hwan (2019)~\cite{sellami2019robust}, Lopez et al. (2020)~\cite{lopez2020artificial}, and Gu and Cai (2021)~\cite{gu2021fusing} utilized algorithm level solutions to handle imbalanced data problems. From this observation, it was also understood that many recent researchers prefer algorithm-level solutions instead of data-level solutions due to their low time complexity issues. Among all of the existing data level solutions, Synthetic Minority Over-Sampling Techniques (SMOTE) is most widely used, as shown in Fig.~\ref{fig:fig12}. For instance, Krishnan et al. (2021) proposed Recurrent Neural Network (RNN) and LSTM-based approaches to detect heart patients using the Cleveland dataset, where SMOTE is applied to balance the data, and the overall model accuracy was around 98.5\%~\cite{krishnan2021hybrid}. Rai and Chatterjee (2021) used SMOTE-TomekLink along with CNN and LSTM to detect MI~\cite{rai2021hybrid}. Apart from SMOTE, other data level approaches such as Under-Sampling~\cite{rezaei2021novel} and Random Over-Sampling~\cite{shah2020heart} is also adopted by few studies.
\\
Among all of the algorithm-level solutions, DL-based solutions are widely utilized by the researchers~\cite{li2020craftnet,romdhane2020electrocardiogram,gu2021fusing}. For instance, Lu et al. (2021) developed a CNN model named Depth wise separable CNN with focal loss (DSC-FL-CNN) to detect arrhythmia~\cite{lu2021automated}. The focal loss (FL) is used to deal with the imbalanced ECG data. Using the MIT-BIH dataset, the authors show that the proposed model can reach an overall F1-score of 0.79 for arrhythmia classification. On the other hand, some study uses the GAN model to augment the ECG data to reduce data scarcity~\cite{wang2021cab,rath2021exhaustive,puspitasari2021generative}. For example, Wang et al. (2021) utilized a Gan-based approach to handle the imbalance problem and achieved an accuracy of 99.71\% in detecting arrhythmia patients~\cite{wang2021cab}. Puspitasari et al. (2021) proposed a Time Series Generative Adversarial Networks (TSGAN) to analyze the fetal heart rate (FHR) signal data. Their preliminary computational results show that GAN-based approaches increased the quality index of 3\%-44\% from other models~\cite{puspitasari2021generative}.\\
Apart from algorithm and data level solutions, few studies also considered cost-sensitive approaches to handle the imbalanced data problem associated with the heart disease-related dataset~\cite{daraei2017efficient,gan2020integrating}. Daraei and Hamidi (2017) applied the Metacost classifier to the imbalance dataset. Metacost made a cost sensitive J48 model by assigning different costs ratios for misclassified cases; including 1:10, 1:50, 1:100, 1:150, and 1:200. With a cost ratio of 1:200, the model outperformed other models that did not use feature selection. The model has a sensitivity of 86.67\%, an F1-score of 80\%, and an accuracy of 82.67\%~\cite{daraei2017efficient}. Gan et al. (2020) suggest that a cost-sensitive approach is more effective than searching for an excellent classifier to maximize classification accuracy. In this regard, the authors proposed an integrated TANBN with a cost-sensitive classification algorithm (AdaC-TANBN) to handle the class imbalance problem with the Cleveland dataset~\cite{gan2020integrating}.\\
The approaches to handle the imbalanced data problems addressed in this review are quite a few among many other existing techniques. However, all solutions are still just the improvised version of data level, algorithm level, or cost-sensitive levels solutions. 
\section{Discussions}\label{discussion}
To get the insights of current trends and techniques used in heart disease diagnosis using an imbalanced dataset, in-depth analysis is carried out using 49 referenced literature. The overall in-depth analysis considered the following factors: heart disease type, applications, ML algorithms, and imbalance solution.\\
Based on the overall analysis, it can be assumed that arrhythmia is the most widely studied heart disease in terms of Machine Learning-based heart disease diagnosis (MLBHDD). At the same time, some other studies also considered cardiac arrest~\cite{liu2012intelligent}, Myocardial infarction (MI)~\cite{daraei2017efficient}, and so on. Note that, Cardiac arrest is still one of the major issues in intensive care units with low survival rates. It is often challenging to screen accurately using DL and traditional ML approaches due to the low sensitivity and high false alarm rates~\cite{baral2021novel}. Therefore, researchers and practitioners need to draw attention to all types of heart disease instead of emphasizing arrhythmia. \\
Most of the ML-based model is developed to detect heart disease patients focusing on feature selections, image segmentation, and classification. Most researchers considerd two popular datasets: Cleveland data and MIT-BIH arrhythmia, extensively due to their availability and challenges associated with data imbalance problems. However, few studies considered real-world data~\cite{marateb2018prediction,salman2019heart,che2021constrained} as well.  Therefore, the model's performance variability is observed during the study results presented in terms of open repository data and real-world data. However, it can not be ignored that the models' performance will be more authentic when the experiment will be carried out using actual world data. Therefore, instead of open repository data, it is necessary to use real-world data to observe the ML-based models' performance. One of the main reasons for the limited Clinical Decision Support Systems (CDSS) system is the instability of the model. As clinical systems cannot rely on only old patient data, thus a continuous model adjustment and development is necessary based on new patient data to run the CDSS system smoothly and correctly. Data collected from real-time and training them using the ML model should be challenging in many situations, i.e., emergency operation theater, ICU patient care.\\
Several ML algorithms such as SVM, KNN, ANN, CNN, GAN are being used to develop the MLBHDD model. However, among all of them, the CNN-based model draws more attention to researchers due to its robust performance and capability to handle complex datasets. Additionally, GAN is recently reported by many referenced literature due to its immense capacity to generate fake data that are very close to actual data, ultimately helping to handle the imbalance data problem. \\
The imbalance data ratio problems have been addressed mostly using data level and algorithm solutions. Among all the data level solutions, SMOTE is the most widely used and is still being used by recent referenced literature. Additionally, the CNN-based approach has become popular among researchers as algorithm-level solutions, which are more prominent in recently published literature. However, one of the limitations of the CNN-based approach is that there is no explanation of how the model interprets the final predictions.\\
While many ML-based heart disease models demonstrated better performance under the intra-patient paradigm, the performance of the inter-patient paradigm is still poor~\cite{yang2021ensemble}. Additionally, it is still challenging and time-consuming to distinguish different heartbeats from ECG as they typically contain noise~\cite{romdhane2020electrocardiogram}. Therefore, even though most of the studies claimed their proposed model could perform well with imbalanced data, still the model would be only trusted if the model results is presented with explainable AI.\\
Regarding the issues with imbalanced class, Most conventional classification models just strive for an optimal classifier to maximize classification accuracy with a constant misclassification cost, not taking into account that the misclassification cost might fluctuate with the sample probability distribution of the samples~\cite{gan2020integrating}. Additionally, most of the referenced literature applied multiple steps such as denoising, heartbeat segmentation, feature extractions, and classifications which are often computationally expensive~\cite{romdhane2020electrocardiogram}. Therefore, implementing such a model in the real world would be challenging and might be an exciting topic for future research.
\\
There is no clear common standard process that researchers and practitioners must follow, particularly for researchers from multidisciplinary or non-medical areas. There are no precise requirements for the presentation of the study report. For instance, some studies emphasize their study result based on accuracy~\cite{exarchos2014multiscale,acharya2017deep}, while some research claimed their best model based on AUC, ROC, sensitivity, and specificity~\cite{daraei2017efficient,marateb2018prediction,awan2019machine,lopez2020artificial}.  Despite the fact that all of those performance measuring tools are utilized to display the statistical results of the majority of the ML classifier's performance, a correct guideline is still necessary to report the study's results effectively. In this regard, a new study~\cite{schwendicke2021better} provides some valuable advice.\\
Nearly all ML algorithms, particularly Deep Learning approaches, are still under investigation on how the model arrived at its final predictions. Even though various explainable or interpretable methodologies have recently been proposed, the ML-based model still fails to provide improved explanations for the model's trustworthiness and interpretation~\cite{brunese2020explainable,ahsan2021detection,fletcher2021addressing}. Unfortunately, among the pooled 451 literature none of them provided any explanation regarding how their model is trustable in terms of predictions. Explanations for ML algorithms are vital, especially in clinical diagnosis systems, because the model must often decide where the patient is in critical conditions such as life and death situations~\cite{ahsan2020deep}. It is also not surprising that many doctors, nurse practitioners, and non-experts may be in charge of those apps, which are dangerous to use in the clinical setting without adequate model interpretation.\\
Secure diagnosis is one of the significant concerns of ML-based clinical diagnosis systems. Geological location, data sample, and heart disease types are some of the elements that could affect the model's effectiveness. For example, ML models developed to diagnose arrhythmia disease may not function for the diagnosis of cardiac arrest. As a result, establishing an ML-based clinical diagnosis system for each disease may necessitate distinct models, as the training data will differ from one data source to the next. Therefore, creating a model that can provide multiple disease diagnoses in real-time must be challenging. Another important aspect of secure diagnosis is the model's stability when parameters change, or models are updated depending on the user's experience.\\
The overall findings of the referenced 49 literature are combined in Table~\ref{tab:tab6} in order to provide a better insights of ML based heart disease diagnosis.
\section{Conclusions}\label{conclusions}
This study aims to identify the ML-based and data-driven recent trends and techniques in heart disease diagnosis with imbalanced data. The post-analysis of this study is that to develop ML-based heart disease diagnosis systems in the real world, and it is necessary to expand the ML-based experiments that include real-time patients’ data and proper explanation of the final prediction using interpretable machine learning. The in-depth analysis of selected 49 papers indicates that there is a need for future studies that must demonstrate the trustable performance in medical domains. Deep Learning dominates this field, while for imbalanced data handling, SMOTE is still one of the popular Over-Sampling techniques and adapted by many theorists and researchers. The rise of GAN based heart disease diagnosis model is also identified and adapted by many researchers due to its ability to develop synthetic data, even though GAN is computationally still expensive. Future research direction might be aligned with the ML-based limitations addressed in the discussion sections. In any case, we expect that ML-based heart disease diagnosis with imbalanced data still has unexplored aspects and many potentials to unlock in the coming years.

\setlength{\LTcapwidth}{\textwidth}
\begin{longtable}
{@{}p{.08\linewidth}p{.1\linewidth}p{.2\linewidth}p{.1\linewidth}p{.1\linewidth}@{}p{.1\linewidth}p{.15\linewidth}p{.1\linewidth}@{}}
\caption{Literature emphasized on Machine Learning based heart disease diagnosis using imbalanced data. CNN-- Convolutional Neural Network; DNN-- Deep Neural network; ANN -- Artificial Neural Network; $S_{e}$-- Sensitivity; $S_{p}$-- Specificity;$A_{c}$-- Accuracy; $P_{c}$-- Precision; $R_{c}$-- Recall; $F_{c}$-- F1-Score; PPV-- Predictive Positive Value; NPV-- Negative Predictive Value; TPR-- True Positive Rate; TNR-- True Negative Rate}\label{tab:tab6}\\
\toprule
  Author(s)&	Heart disease type&	Application&	ML-Algorithms&	Imbalance solutions&	Approach&	Evaluations&	Data \\\midrule\endfirsthead
       \caption{\emph{Cont.}}\\\toprule
       Author(s)&	Heart disease type&	Application&	ML-Algorithms&	Imbalance solutions&	Approach&	Evaluations&	Data\\\midrule
       \endhead
       
     Liu et al. (2012)~\cite{liu2012intelligent}&	Cardiac arrest&	Intelligent scoring system for the prediction of cardiac arrest within 72h&	SVM

	&Data level&	Under sampling	&$S_{e}$-78.8\%,$S_{p}$-62.3\%,PPV-10\%,NPV-98.2\% & 1386 patients data\\
Sree et al. (2012)~\cite{sree2012cardiac}&	Cardiac arrhythmias&	Analyzed HRV signal abnormalities to determine and classify arrhythmias&	Probabilistic neural network (PNN)&	&		&$A_{c}$-80\%,
$S_{e}$-82\%,$S_{p}$-85.6\%&90 patients data\\
Nahar et al. (2013)~\cite{nahar2013computational}&	Heart disease (general)&	Investigates a number of computational intelligence techniques in the detection of heart disease&	NB, IBK and SMO&	Data level&	Medical knowledge driven feature selection process (MFS)&	$A_{c}$- 83\%-97.05\% &	Cleveland dataset\\
Exarchos et al. (2015)~\cite{exarchos2014multiscale}& Heart disease (general)& Model the progression of atherosclerosis (ATS)&	Bayesian networks&	Data level&	SMOTE&	$A_{c}$-83\%-93\%&	39 patients data\\
Bashir et al. (2016)~\cite{bashir2016multicriteria}&	Heart disease (general)&	Focuses on prediction and analysis of heart disease (general)&	Ensemble approach(NB, DT based Gini index, DT information gain, instance-based learner, and SVM)&	Data level&	Tenfold cross-validation&	$A_{c}$-87.37\%, $S_{e}$-93.75\%, $S_{p}$-92.86\%, F-measure -82.17\%	&UCI repository dataset\\
Wilharto et al. (2016)~\cite{wiharto2016intelligence}&	Coronary heart disease&
An intelligence systems to diagnosis coronary heart disease&	K-star algorithms&	Data level&	SMOTE&	$S_{e}$-80.1\%, $S_{p}$-95\%, $F_{c}$- 80.1\%&	Cleveland dataset\\
Acharya et al. (2017)~\cite{acharya2017deep}&	Arrhythmia diagnosis&Automatically identify 5 different categories of heartbeats&	CNN&	Data level&	Data augmentation&	$A_{c}$-94\% (balance data),
$A_{c}$-89.07\% (imbalance data)&	MIT-BIH\\  
 Daraei and Hamidi (2017)~\cite{daraei2017efficient}&	Myocardial infarction (MI)&MI prediction model data mining&J48&	Cost sensitive&	Metacost&	$S_{e}$- 86.7\%, $F_{c}$- 80\%, $A_{c}$- 82.57\%&	455 healthy and 295 myocardial infarction cases\\

Yang et al. (2018)~\cite{yang2018automatic}&	Arrhythmia& Heartbeat recognition method is presented&	SVM&	Algorithm level&	Principal component analysis network (PCANet)&	$A_{c}$- 97.77\% (imbalance data),
$A_{c}$- 97.08\%( noise-free ECGs)&	MIT-BIH\\
Polat et al. (2018)~\cite{polat2018similarity}&	Heart disease (general)&	A new data preprocessing method&
	LDA, KNN, SVM, and RF&	Data level&	Random sub-sampling&	$A_{c}$- 84\%-97\%&	UCI machine learning repository\\

Rajesh and Dhuli (2018)~\cite{rajesh2018classification}&	Arrhythmia&	Predict the presence of arrhythmia&
	AdaBoost ensemble  classifier&	Data level&	Re-sampling,
SMOTE, Distribution based data sampling&	$S_{e}$ - 96.5\%,
$S_{p}$- 99.1\%, ROC-99.5\%, $A_{c}$- 98.6\%&	MIT-BIH\\
Sellami and Hwan (2019)~\cite{sellami2019robust}&	Abnormal heart rhythm&	Heartbeat classification&	CNN&	Algorithm level&	Batch-weighted loss function&	$A_{c}$- 99.48\%, PPV- 98.83\%,
$S_{e}$- 96.97\%, $S_{p}$-99.87\%&	MIT-BIH\\
Awan et al. (2019)~\cite{awan2019machine}&	Heart failure&	Predict 30 day HF readmission or death&	MLP&	Algorithm level&	MLP-based approaches&	AUC- 0.62, $S_{e}$- 48\%, $S_{p}$- 70\%&	Western Australian patients (2003-2008)\\
Salman (2019)~\cite{salman2019heart}&	Acute myocardial infarction&	Hospital mortality for patients&	CNN&	Algorithm level&	Chow–Liu and tree-augmented naive Bayesian&	$A_{c}$- 99.79\%&	Real data of about 787 patients\\
Lopez et al. (2020)~\cite{lopez2020artificial}&	Hypertensive&
	Estimate the association among gender, race, BMI, age, smoking, kidney disease and diabetes in hypertensive patients&	ANN&	Data level&	SMOTE&	$S_{e}$- 40\%, $S_{p}$- 87\%, $P_{c}$- 57.8\% and AUC- 0.77&	National health and nutrition examination survey (2007-2016)\\
	Dutta et al. (2020)~\cite{dutta2020efficient}&	Coronary Heart Disease&	Classify significantly class-imbalanced clinical data&	CNN&	Algorithm level&	&	$A_{c}$-79.5\%	&National health and nutritional examination survey (NHANES)\\
Ebiaredoh and Esenogho (2020)~\cite{ebiaredoh2020integrating}&	Heart disease (general)& A new model for predicting chronic kidney disease, cervical cancer, and heart disease&	Enhanced sparse autoencoder (SAE) and Softmax regression&	Algorithm level&	&	$A_{c}$- 91\%&	Framingham Heart Study Dataset\\
Li et al. (2020)~\cite{li2020craftnet}&	Cardiovascular diseases&	Accurately recognizing the handcraft features&	DNN& CraftNet&	Algorithm level&	$A_{c}$– (86.82\%-89.25\%)&	MIT-BIH\\

Romdhane et al. (2020)~\cite{romdhane2020electrocardiogram}&	Arrhythmia detection&	An algorithm for heartbeat segmentation&	CNN&	Algorithm level&	Focal loss&	$A_{c}$- 98.41\%,
$F_{c}$- 98.38\%, $P_{c}$- 98.37\%, $R_{c}$- 98.41\%&	MIT-BIH and INCART\\      
 Wang et al. (2020)~\cite{wang2020left}&	Heart disease (general)&	Accurate left ventricle landmark localization&	CNN&	Algorithm level&	Focal loss&	$A_{c}$- 95.82\%&	Cardiac atlas project (CAP) data set\\
         Minou et al. (2020)~\cite{minou2020classification}&	
Cardiovascular diseases&	Build and compare classification techniques for CVD&
	RF, DT&	Data level&	SMOTE&	$A_{c}$- 68\%-91\%,
$P_{c}$-0-79\%,
$R_{c}$-0-82.6\%,
$F_{c}$-0-81\%&	4270 patients data\\
Gan et al. (2020)~\cite{gan2020integrating}&	Heart disease&	Overcome the imbalance problem&	AdaC-TANBN&	Cost sensitive&	TANBN with cost-sensitive classification (AdaC-TANBN)&	$A_{c}$- 0.80$\pm$0.371,AUC- 0.8887$\pm$0.0268&	Cleveland dataset\\
Sharma and Sunkaria (2020)~\cite{sharma2020myocardial}& 	Myocardial Infarction&	Present a technique for MI detection and localization&	KNN &	Data level&	Adaptive synthetic (ADASYN)	$A_{c}$- 99.76\%&	Physikalisch-Technische Bundesanstalt (PTB) database\\
Shah et al. (2020)~\cite{shah2020heart}&	Heart disease (general)&	Predict heart disease at the earliest, help prioritize hospital consultations& SVM, RF, Ordinal Regression, LR and NB&	Data level&	Sampling&	$A_{c}$- 95\% (using SVM)&	Cleveland dataset\\
Wang et al. (2020)~\cite{wang2020multi}&	Heart failure&	Mortality rate of patient&	Multi-view ensemble learning algorithm based on empirical kernel mapping( MVE-EK)&
	Data level&	Under-Sampling&	TPR- 81.07\%, TNR- 77.35\%, AUC-89.64\%&	Shanghai Shuguang Hospital data\\
Krishnan et al. (2021)~\cite{krishnan2021hybrid}&	Heart disease (general)&	Heart disease prediction&	RNN+ LSTM&	Data level&	SMOTE&	$A_{c}$- 98.5\%&	Cleveland dataset\\
Plati et al. (2021)~\cite{plati2021machine}&	Chronic heart failure (HF)&	Investigated the incremental value of each feature&
	DT, RF, rotation forest (ROT), NB, KNN, SVM, logistic model tree (LMT), and Bayes network (BN)& 	Data level&	Under-Sampling&	$A_{c}$ -91.23\%, $S_{e}$– 93.83\%, $S_{p}$– 89.62\%&	487 patient data   \\
Baral et al. (2021)~\cite{baral2021novel}&	Cardiac arrest&	Early prediction of cardiac arrest in Sepsis patient&	MLP and enhanced Bidirectional LSTM&	Data level&	SMOTE&	Accuracy-92.6\%, $S_{c}$– 94.3\%,
$S_{p}$- 93.6\%, AUC- 0.94&	Medical information mart for intensive care (MIMIC-III) database\\
Dixit and Kala (2021)~\cite{dixit2021early}&	Heart diseases&	Early detection of heart diseases using a low-cost compact ECG sensor&	1D CNN&	Data level&	Over-Sampling&	$A_{c}$- 93\%& 300 participants data\\
Lu et al. (2021)~\cite{lu2021automated}&	Arrhythmia& 	Automated arrhythmia classification&	Depth wise separable convolutional neural network with focal loss (DSC-FL-CNN)&	Algorithm level&	Focal loss&	$F_{c}$- 0.79&	MIT-BIH\\
 Wang et al. (2021)~\cite{wang2021cab}&	Arrhythmia &	Classifying arrhythmia using imbalanced data&	GAN-based deep learning framework (called CAB)&	Data level&	Data augmentation using GAN&	$A_{c}$- 99.71\%&	MIT-BIH\\      
Rezaei et al. (2021)~\cite{rezaei2021novel}&	Heart arrhythmia&	Heart arrhythmia detection with imbalance data&	XGBoost Classifier&	Data level&	Under-Sampling&	$F_{c}$- 87.22\%, $S_{e}$- 88.55\%, $S_{p}$- 85.95\%&	UK Biobank dataset\\
Ammar et al. (2021)~\cite{ammar2021automatic}&	Heart disease&	Automatic cardiac cine MRI segmentation&	CNN&
	Data level&	Data augmentation&	$A_{c}$- 92\%&	150 patients data from Dijon Hospital\\
Zhu et al. (2021)~\cite{zhu2021segmentation}&	Coronary heart disease&

	Segmentation of Coronary Arteries Images&	Spatio-temporal feature fusion network with combo loss&

	Algorithm level&	Loss function&	$A_{c}$- 87\%,	Coronary artery CTA image data\\
Rai and Chatterjee (2021)~\cite{rai2021hybrid}
&	MI&	Automatic and accurate prognosis of MI using ECG&
	Hybrid CNN-LSTM&	Data level&	SMOTE-Tomek Link&	$A_{c}$- 99.88\%&	MIT-BIH\\
	Puspitasari et al. (2021)~\cite{puspitasari2021generative}&	Fetal heart rate&	Unbalanced fetal heart rate signal classification&	Time series generative adversarial networks (TSGAN)&	Data level&	Data augmentation&	$A_{c}$- 71.08\%, $S_{e}$- 67.64\%, $S_{p}$- 71.97\%, QI-69.77\%&	 CTU– UHB Intrapartum Cardiotocography Database\\
Ketu and Mishra (2021)~\cite{ketu2021empirical}&	Arrhythmia&	Heart disease diagnosis on imbalance dataset& SVM, KNN, RF, ET,DT, LR, and AB&	Data level&	SMOTE&	$A_{c}$- 92\%-99\%&	MIT-BIH\\	
        Khdair and Dasari (2021)~\cite{khdair2021exploring}&	Coronary Heart Disease&	Predict the occurrence of CHD events from clinical data&	LR, SVM, KNN, and MLP-NN&	Data level&	SMOTE&	$A_{c}$-72.7\%-73.8\%,
$P_{c}$- 63.3\%-70\%,
$R_{c}$- 39.4\%-50.6\%, $F_{c}$- 50.4\%-56.3\%, $S_{p}$- 84.4\%-91.1\%&	South African heart disease data\\
Waqar et al. (2021)~\cite{waqar2021efficient}&	Heart attack&	Predict the heart disease&	ANN&	Data level&	SMOTE&	$A_{c}$- 96\%, $P_{c}$-96.1\%, $R_{c}$-95.7\%, $F_{c}$-
95.7\%&	UCI dataset\\
Sharmila (2021)~\cite{sharmilamulti}&	Arrhythmia&
	Multi-Class Arrhythmia Detection using a hybrid spatial-temporal feature extraction&	DNN&

	Butter worth filter&	Algorithm level&	$A_{c}$-99.65\%,$P_{c}$- 97.15\%,$S_{e}$- 99.34\%,$S_{p}$- 99.82\%,$F_{c}$- 96.59\%&	MIT-BIH\\

Yang et al. (2021)~\cite{yang2021ensemble}&	Arrhythmia&
	Automatic heartbeat classification&
	Ensemble learning and multi-kernel learning	baseline removal&	Algorithm level&	$A_{c}$- 98.1\%&	MIT-BIH\\
    \bottomrule
    \label{ref:big table}
\end{longtable}

\bibliographystyle{unsrt}  
\bibliography{main}

\begin{thebibliography}{10}

\bibitem{ahsan2021effect}
Md~Manjurul Ahsan, MA~Mahmud, Pritom~Kumar Saha, Kishor~Datta Gupta, and Zahed
  Siddique.
\newblock Effect of data scaling methods on machine learning algorithms and
  model performance.
\newblock {\em Technologies}, 9(3):52, 2021.

\bibitem{WHO}
Cardiovascular disease.
\newblock
  \url{https://www.who.int/health-topics/cardiovascular-diseases#tab=tab_1},
  November 25, 2021.

\bibitem{lip2017antithrombotic}
Gregory~YH Lip, Jean~Philippe Collet, Raffaele~de Caterina, Laurent Fauchier,
  Deirdre~A Lane, Torben~B Larsen, Francisco Marin, Joao Morais, Calambur
  Narasimhan, Brian Olshansky, et~al.
\newblock Antithrombotic therapy in atrial fibrillation associated with
  valvular heart disease: a joint consensus document from the european heart
  rhythm association (ehra) and european society of cardiology working group on
  thrombosis, endorsed by the esc working group on valvular heart disease,
  cardiac arrhythmia society of southern africa (cassa), heart rhythm society
  (hrs), asia pacific heart rhythm society (aphrs), south african heart (sa
  heart) association and sociedad latinoamericana de estimulacion cardiaca y
  electrofisiologia (soleace).
\newblock {\em Ep Europace}, 19(11):1757--1758, 2017.

\bibitem{islam2013coronary}
AKM~Monwarul Islam and AAS Majumder.
\newblock Coronary artery disease in bangladesh: A review.
\newblock {\em Indian heart journal}, 65(4):424--435, 2013.

\bibitem{nahar2013computational}
Jesmin Nahar, Tasadduq Imam, Kevin~S Tickle, and Yi-Ping~Phoebe Chen.
\newblock Computational intelligence for heart disease diagnosis: A medical
  knowledge driven approach.
\newblock {\em Expert Systems with Applications}, 40(1):96--104, 2013.

\bibitem{sree2012cardiac}
S~Vinitha Sree, Dhanjoo~N Ghista, and Kwan-Hoong Ng.
\newblock Cardiac arrhythmia diagnosis by hrv signal processing using principal
  component analysis.
\newblock {\em Journal of Mechanics in Medicine and Biology}, 12(05):1240032,
  2012.

\bibitem{liu2012intelligent}
Nan Liu, Zhiping Lin, Jiuwen Cao, Zhixiong Koh, Tongtong Zhang, Guang-Bin
  Huang, Wee Ser, and Marcus Eng~Hock Ong.
\newblock An intelligent scoring system and its application to cardiac arrest
  prediction.
\newblock {\em IEEE Transactions on Information Technology in Biomedicine},
  16(6):1324--1331, 2012.

\bibitem{exarchos2014multiscale}
Konstantinos~P Exarchos, Clara Carpegianni, Georgios Rigas, Themis~P Exarchos,
  Federico Vozzi, Antonis Sakellarios, Paolo Marraccini, Katerina Naka, Lambros
  Michalis, Oberdan Parodi, et~al.
\newblock A multiscale approach for modeling atherosclerosis progression.
\newblock {\em IEEE journal of biomedical and health informatics},
  19(2):709--719, 2014.

\bibitem{wiharto2016intelligence}
Wiharto Wiharto, Hari Kusnanto, and Herianto Herianto.
\newblock Intelligence system for diagnosis level of coronary heart disease
  with k-star algorithm.
\newblock {\em Healthcare informatics research}, 22(1):30--38, 2016.

\bibitem{bashir2016multicriteria}
Saba Bashir, Usman Qamar, and Farhan~Hassan Khan.
\newblock A multicriteria weighted vote-based classifier ensemble for heart
  disease prediction.
\newblock {\em Computational Intelligence}, 32(4):615--645, 2016.

\bibitem{daraei2017efficient}
Atefeh Daraei and Hodjat Hamidi.
\newblock An efficient predictive model for myocardial infarction using
  cost-sensitive j48 model.
\newblock {\em Iranian journal of public health}, 46(5):682, 2017.

\bibitem{dutta2020efficient}
Aniruddha Dutta, Tamal Batabyal, Meheli Basu, and Scott~T Acton.
\newblock An efficient convolutional neural network for coronary heart disease
  prediction.
\newblock {\em Expert Systems with Applications}, 159:113408, 2020.

\bibitem{li2020craftnet}
Yong Li, Zihang He, Heng Wang, Bohan Li, Fengnan Li, Ying Gao, and Xiang Ye.
\newblock Craftnet: a deep learning ensemble to diagnose cardiovascular
  diseases.
\newblock {\em Biomedical Signal Processing and Control}, 62:102091, 2020.

\bibitem{ahsan2021detection}
Md~Manjurul Ahsan, Redwan Nazim, Zahed Siddique, and Pedro Huebner.
\newblock Detection of covid-19 patients from ct scan and chest x-ray data
  using modified mobilenetv2 and lime.
\newblock In {\em Healthcare}, volume~9, page 1099. Multidisciplinary Digital
  Publishing Institute, 2021.

\bibitem{ahsan2021detecting}
Md~Manjurul Ahsan, Md~Tanvir Ahad, Farzana~Akter Soma, Shuva Paul, Ananna
  Chowdhury, Shahana~Akter Luna, Munshi Md~Shafwat Yazdan, Akhlaqur Rahman,
  Zahed Siddique, and Pedro Huebner.
\newblock Detecting sars-cov-2 from chest x-ray using artificial intelligence.
\newblock {\em Ieee Access}, 9:35501--35513, 2021.

\bibitem{fernandez2011addressing}
Alberto Fern{\'a}ndez, Salvador Garc{\'\i}a, and Francisco Herrera.
\newblock Addressing the classification with imbalanced data: open problems and
  new challenges on class distribution.
\newblock In {\em International conference on hybrid artificial intelligence
  systems}, pages 1--10. Springer, 2011.

\bibitem{benhar2020data}
Houda Benhar, Ali Idri, and JL~Fernandez-Aleman.
\newblock Data preprocessing for heart disease classification: A systematic
  literature review.
\newblock {\em Computer Methods and Programs in Biomedicine}, page 105635,
  2020.

\bibitem{rath2021exhaustive}
Adyasha Rath, Debahuti Mishra, Ganapati Panda, and Suresh~Chandra Satapathy.
\newblock An exhaustive review of machine and deep learning based diagnosis of
  heart diseases.
\newblock {\em Multimedia Tools and Applications}, pages 1--59, 2021.

\bibitem{hoodbhoy2021diagnostic}
Zahra Hoodbhoy, Uswa Jiwani, Saima Sattar, Rehana Salam, Babar Hasan, and Jai~K
  Das.
\newblock Diagnostic accuracy of machine learning models to identify congenital
  heart disease: A meta-analysis.
\newblock {\em Frontiers in artificial intelligence}, 4:97, 2021.

\bibitem{verma2021effective}
Simran Verma and Abhishek Gupta.
\newblock Effective prediction of heart disease using data mining and machine
  learning: A review.
\newblock In {\em 2021 International Conference on Artificial Intelligence and
  Smart Systems (ICAIS)}, pages 249--253. IEEE, 2021.

\bibitem{kumar2021machine}
Narender Kumar and Dharmender Kumar.
\newblock Machine learning based heart disease diagnosis using non-invasive
  methods: A review.
\newblock In {\em Journal of Physics: Conference Series}, volume 1950, page
  012081. IOP Publishing, 2021.

\bibitem{okoli2010guide}
Chitu Okoli and Kira Schabram.
\newblock A guide to conducting a systematic literature review of information
  systems research.
\newblock {\em Sprouts}, 2010.

\bibitem{tricco2018prisma}
Andrea~C Tricco, Erin Lillie, Wasifa Zarin, Kelly~K O'Brien, Heather Colquhoun,
  Danielle Levac, David Moher, Micah~DJ Peters, Tanya Horsley, Laura Weeks,
  et~al.
\newblock Prisma extension for scoping reviews (prisma-scr): checklist and
  explanation.
\newblock {\em Annals of internal medicine}, 169(7):467--473, 2018.

\bibitem{fahimnia2015green}
Behnam Fahimnia, Joseph Sarkis, and Hoda Davarzani.
\newblock Green supply chain management: A review and bibliometric analysis.
\newblock {\em International Journal of Production Economics}, 162:101--114,
  2015.

\bibitem{malviya2015green}
Rakesh~Kumar Malviya and Ravi Kant.
\newblock Green supply chain management (gscm): a structured literature review
  and research implications.
\newblock {\em Benchmarking: An international journal}, 2015.

\bibitem{acharya2017deep}
U~Rajendra Acharya, Shu~Lih Oh, Yuki Hagiwara, Jen~Hong Tan, Muhammad Adam,
  Arkadiusz Gertych, and Ru~San~Tan.
\newblock A deep convolutional neural network model to classify heartbeats.
\newblock {\em Computers in biology and medicine}, 89:389--396, 2017.

\bibitem{plawiak2020novel}
Pawe{\l} P{\l}awiak and U~Rajendra Acharya.
\newblock Novel deep genetic ensemble of classifiers for arrhythmia detection
  using ecg signals.
\newblock {\em Neural Computing and Applications}, 32(15):11137--11161, 2020.

\bibitem{novikov2018fully}
Alexey~A Novikov, Dimitrios Lenis, David Major, Jivri Hladuuvka, Maria Wimmer,
  and Katja Buhler.
\newblock Fully convolutional architectures for multiclass segmentation in
  chest radiographs.
\newblock {\em IEEE transactions on medical imaging}, 37(8):1865--1876, 2018.

\bibitem{rajesh2018classification}
Kandala~NVPS Rajesh and Ravindra Dhuli.
\newblock Classification of imbalanced ecg beats using re-sampling techniques
  and adaboost ensemble classifier.
\newblock {\em Biomedical Signal Processing and Control}, 41:242--254, 2018.

\bibitem{yang2018automatic}
Weiyi Yang, Yujuan Si, Di~Wang, and Buhao Guo.
\newblock Automatic recognition of arrhythmia based on principal component
  analysis network and linear support vector machine.
\newblock {\em Computers in biology and medicine}, 101:22--32, 2018.

\bibitem{sellami2019robust}
Ali Sellami and Heasoo Hwang.
\newblock A robust deep convolutional neural network with batch-weighted loss
  for heartbeat classification.
\newblock {\em Expert Systems with Applications}, 122:75--84, 2019.

\bibitem{awad2017early}
Aya Awad, Mohamed Bader-El-Den, James McNicholas, and Jim Briggs.
\newblock Early hospital mortality prediction of intensive care unit patients
  using an ensemble learning approach.
\newblock {\em International journal of medical informatics}, 108:185--195,
  2017.

\bibitem{sakr2017comparison}
Sherif Sakr, Radwa Elshawi, Amjad~M Ahmed, Waqas~T Qureshi, Clinton~A Brawner,
  Steven~J Keteyian, Michael~J Blaha, and Mouaz~H Al-Mallah.
\newblock Comparison of machine learning techniques to predict all-cause
  mortality using fitness data: the henry ford exercise testing (fit) project.
\newblock {\em BMC medical informatics and decision making}, 17(1):1--15, 2017.

\bibitem{awan2019machine}
Saqib~Ejaz Awan, Mohammed Bennamoun, Ferdous Sohel, Frank~Mario Sanfilippo, and
  Girish Dwivedi.
\newblock Machine learning-based prediction of heart failure readmission or
  death: implications of choosing the right model and the right metrics.
\newblock {\em ESC heart failure}, 6(2):428--435, 2019.

\bibitem{romdhane2020electrocardiogram}
Taissir~Fekih Romdhane and Mohamed~Atri Pr.
\newblock Electrocardiogram heartbeat classification based on a deep
  convolutional neural network and focal loss.
\newblock {\em Computers in Biology and Medicine}, 123:103866, 2020.

\bibitem{che2021constrained}
Chao Che, Peiliang Zhang, Min Zhu, Yue Qu, and Bo~Jin.
\newblock Constrained transformer network for ecg signal processing and
  arrhythmia classification.
\newblock {\em BMC Medical Informatics and Decision Making}, 21(1):1--13, 2021.

\bibitem{gan2020integrating}
Dan Gan, Jiang Shen, Bang An, Man Xu, and Na~Liu.
\newblock Integrating tanbn with cost sensitive classification algorithm for
  imbalanced data in medical diagnosis.
\newblock {\em Computers \& Industrial Engineering}, 140:106266, 2020.

\bibitem{minou2020classification}
John Minou, John Mantas, Flora Malamateniou, and Daphne Kaitelidou.
\newblock Classification techniques for cardio-vascular diseases using
  supervised machine learning.
\newblock {\em Medical Archives}, 74(1):39, 2020.

\bibitem{kumar2021cardiovascular}
Mikkili~Dileep Kumar and KV~Ramana.
\newblock Cardiovascular disease prognosis and severity analysis using hybrid
  heuristic methods.
\newblock {\em Multimedia Tools and Applications}, 80(5):7939--7965, 2021.

\bibitem{baral2021novel}
Samit Baral, Abeer Alsadoon, PWC Prasad, Sarmad Al~Aloussi, and Omar~Hisham
  Alsadoon.
\newblock A novel solution of using deep learning for early prediction cardiac
  arrest in sepsis patient: enhanced bidirectional long short-term memory
  (lstm).
\newblock {\em Multimedia Tools and Applications}, 80(21):32639--32664, 2021.

\bibitem{sharma2020myocardial}
LD~Sharma and RK~Sunkaria.
\newblock Myocardial infarction detection and localization using optimal
  features based lead specific approach.
\newblock {\em IRBM}, 41(1):58--70, 2020.

\bibitem{dixit2021early}
Shivam Dixit and Rahul Kala.
\newblock Early detection of heart diseases using a low-cost compact ecg
  sensor.
\newblock {\em Multimedia Tools and Applications}, 80(21):32615--32637, 2021.

\bibitem{shah2020heart}
Devansh Shah, Samir Patel, and Santosh~Kumar Bharti.
\newblock Heart disease prediction using machine learning techniques.
\newblock {\em SN Computer Science}, 1(6):1--6, 2020.

\bibitem{wang2020left}
Xuchu Wang, Suiqiang Zhai, and Yanmin Niu.
\newblock Left ventricle landmark localization and identification in cardiac
  mri by deep metric learning-assisted cnn regression.
\newblock {\em Neurocomputing}, 399:153--170, 2020.

\bibitem{polat2018similarity}
Kemal Polat.
\newblock Similarity-based attribute weighting methods via clustering
  algorithms in the classification of imbalanced medical datasets.
\newblock {\em Neural Computing and Applications}, 30(3):987--1013, 2018.

\bibitem{wang2021cab}
Yilin Wang, Le~Sun, and Sudha Subramani.
\newblock Cab: Classifying arrhythmias based on imbalanced sensor data.
\newblock {\em KSII Transactions on Internet and Information Systems (TIIS)},
  15(7):2304--2320, 2021.

\bibitem{puspitasari2021generative}
Riskyana Dewi~Intan Puspitasari, M~Anwar Ma’sum, Machmud~R Alhamidi, Wisnu
  Jatmiko, et~al.
\newblock Generative adversarial networks for unbalanced fetal heart rate
  signal classification.
\newblock {\em ICT Express}, 2021.

\bibitem{lopez2020artificial}
Fernando L{\'o}pez-Mart{\'\i}nez, Edward~Rolando N{\'u}{\~n}ez-Valdez,
  Rub{\'e}n~Gonz{\'a}lez Crespo, and Vicente Garc{\'\i}a-D{\'\i}az.
\newblock An artificial neural network approach for predicting hypertension
  using nhanes data.
\newblock {\em Scientific Reports}, 10(1):1--14, 2020.

\bibitem{gu2021fusing}
Linyan Gu and Xiao-Chuan Cai.
\newblock Fusing 2d and 3d convolutional neural networks for the segmentation
  of aorta and coronary arteries from ct images.
\newblock {\em Artificial Intelligence in Medicine}, 121:102189, 2021.

\bibitem{krishnan2021hybrid}
Surenthiran Krishnan, Pritheega Magalingam, and Roslina Ibrahim.
\newblock Hybrid deep learning model using recurrent neural network and gated
  recurrent unit for heart disease prediction.
\newblock {\em International Journal of Electrical \& Computer Engineering
  (2088-8708)}, 11(6), 2021.

\bibitem{rai2021hybrid}
Hari~Mohan Rai and Kalyan Chatterjee.
\newblock Hybrid cnn-lstm deep learning model and ensemble technique for
  automatic detection of myocardial infarction using big ecg data.
\newblock {\em Applied Intelligence}, pages 1--19, 2021.

\bibitem{rezaei2021novel}
Mercedeh~J Rezaei, John~R Woodward, Julia Ram{\'\i}rez, and Patricia Munroe.
\newblock A novel two-stage heart arrhythmia ensemble classifier.
\newblock {\em Computers}, 10(5):60, 2021.

\bibitem{lu2021automated}
Yi~Lu, Mingfeng Jiang, Liying Wei, Jucheng Zhang, Zhikang Wang, Bo~Wei, and
  Ling Xia.
\newblock Automated arrhythmia classification using depthwise separable
  convolutional neural network with focal loss.
\newblock {\em Biomedical Signal Processing and Control}, 69:102843, 2021.

\bibitem{marateb2018prediction}
Hamid~R Marateb, Mohammad~Reza Mohebian, Shaghayegh~Haghjooy Javanmard,
  Amir~Ali Tavallaei, Mohammad~Hasan Tajadini, Motahar Heidari-Beni,
  Miguel~Angel Ma{\~n}anas, Mohammad~Esmaeil Motlagh, Ramin Heshmat, Marjan
  Mansourian, et~al.
\newblock Prediction of dyslipidemia using gene mutations, family history of
  diseases and anthropometric indicators in children and adolescents: the
  caspian-iii study.
\newblock {\em Computational and structural biotechnology journal},
  16:121--130, 2018.

\bibitem{salman2019heart}
Issam Salman.
\newblock Heart attack mortality prediction: an application of machine learning
  methods.
\newblock {\em Turkish Journal of Electrical Engineering \& Computer Sciences},
  27(6):4378--4389, 2019.

\bibitem{yang2021ensemble}
Ping Yang, Dan Wang, Wen-Bing Zhao, Li-Hua Fu, Jin-Lian Du, and Hang Su.
\newblock Ensemble of kernel extreme learning machine based random forest
  classifiers for automatic heartbeat classification.
\newblock {\em Biomedical Signal Processing and Control}, 63:102138, 2021.

\bibitem{schwendicke2021better}
F~Schwendicke and J~Krois.
\newblock Better reporting of studies on artificial intelligence: Consort-ai
  and beyond.
\newblock {\em Journal of Dental Research}, page 0022034521998337, 2021.

\bibitem{brunese2020explainable}
Luca Brunese, Francesco Mercaldo, Alfonso Reginelli, and Antonella Santone.
\newblock Explainable deep learning for pulmonary disease and coronavirus
  covid-19 detection from x-rays.
\newblock {\em Computer Methods and Programs in Biomedicine}, 196:105608, 2020.

\bibitem{fletcher2021addressing}
Richard~Rib{\'o}n Fletcher, Audace Nakeshimana, and Olusubomi Olubeko.
\newblock Addressing fairness, bias, and appropriate use of artificial
  intelligence and machine learning in global health.
\newblock {\em Frontiers in Artificial Intelligence}, 3:116, 2021.

\bibitem{ahsan2020deep}
Md~Manjurul Ahsan, Tasfiq E~Alam, Theodore Trafalis, and Pedro Huebner.
\newblock Deep mlp-cnn model using mixed-data to distinguish between covid-19
  and non-covid-19 patients.
\newblock {\em Symmetry}, 12(9):1526, 2020.

\bibitem{ebiaredoh2020integrating}
Sarah~A Ebiaredoh-Mienye, Ebenezer Esenogho, and Theo~G Swart.
\newblock Integrating enhanced sparse autoencoder-based artificial neural
  network technique and softmax regression for medical diagnosis.
\newblock {\em Electronics}, 9(11):1963, 2020.

\bibitem{wang2020multi}
Zhe Wang, Lilong Chen, Jing Zhang, Yichao Yin, and Dongdong Li.
\newblock Multi-view ensemble learning with empirical kernel for heart failure
  mortality prediction.
\newblock {\em International journal for numerical methods in biomedical
  engineering}, 36(1):e3273, 2020.

\bibitem{plati2021machine}
Dafni~K Plati, Evanthia~E Tripoliti, Aris Bechlioulis, Aidonis Rammos, Iliada
  Dimou, Lampros Lakkas, Chris Watson, Ken McDonald, Mark Ledwidge, Rebabonye
  Pharithi, et~al.
\newblock A machine learning approach for chronic heart failure diagnosis.
\newblock {\em Diagnostics}, 11(10):1863, 2021.

\bibitem{ammar2021automatic}
Abderazzak Ammar, Omar Bouattane, and Mohamed Youssfi.
\newblock Automatic cardiac cine mri segmentation and heart disease
  classification.
\newblock {\em Computerized Medical Imaging and Graphics}, 88:101864, 2021.

\bibitem{zhu2021segmentation}
Hongyan Zhu, Shuni Song, Lisheng Xu, Along Song, and Benqiang Yang.
\newblock Segmentation of coronary arteries images using spatio-temporal
  feature fusion network with combo loss.
\newblock {\em Cardiovascular Engineering and Technology}, pages 1--12, 2021.

\bibitem{ketu2021empirical}
Shwet Ketu and Pramod~Kumar Mishra.
\newblock Empirical analysis of machine learning algorithms on imbalance
  electrocardiogram based arrhythmia dataset for heart disease detection.
\newblock {\em Arabian Journal for Science and Engineering}, pages 1--23, 2021.

\bibitem{khdair2021exploring}
Hisham Khdair.
\newblock Exploring machine learning techniques for coronary heart disease
  prediction.
\newblock {\em International Journal of Advanced Computer Science and
  Applications}, 12(5), 2021.

\bibitem{waqar2021efficient}
Muhammad Waqar, Hassan Dawood, Hussain Dawood, Nadeem Majeed, Ameen Banjar, and
  Riad Alharbey.
\newblock An efficient smote-based deep learning model for heart attack
  prediction.
\newblock {\em Scientific Programming}, 2021, 2021.

\bibitem{sharmilamulti}
Vallem Sharmila.
\newblock Multi-class arrhythmia detection using a hybrid spatial-temporal
  feature extraction method and stacked auto encoder.
\newblock {\em International Journal of Intelligent Engineering and Systems},
  2020.

\end{thebibliography}

\end{document}